\documentclass[conference]{IEEEtran}
\IEEEoverridecommandlockouts
\usepackage{cite}
\usepackage{amsmath,amssymb,amsfonts}
\usepackage{algorithmic}
\usepackage{graphicx}
\usepackage{textcomp}
\usepackage{xcolor}
\usepackage{balance}

\def\BibTeX{{\rm B\kern-.05em{\sc i\kern-.025em b}\kern-.08em
    T\kern-.1667em\lower.7ex\hbox{E}\kern-.125emX}}
\begin{document}

%\title{Image based Static Facial Expression Recognition in Affect Emotion Recognition via Learning Deep Convolutional Autoencoder *\\

\title{Continuous Emotion Recognition via Deep Convolutional Autoencoder and Support Vector Regressor}
%\\

\author{\IEEEauthorblockN{Sevegni Odilon Clement Allognon \IEEEauthorrefmark{1}, Alceu de S. Britto Jr.\IEEEauthorrefmark{2} and Alessandro L. Koerich \IEEEauthorrefmark{1}}\\ \IEEEauthorblockA{\IEEEauthorrefmark{1}\'Ecole de Technologie Sup\'erieure, Universit\'e du Qu\'ebec, Montr\'eal, QC, Canada\\
Email: sevegni-odilon.allognon.1@ens.etsmtl.ca, alessandro.koerich@etsmtl.ca}
\IEEEauthorblockA{\IEEEauthorrefmark{2}Pontifical Catholic University of Paran\'a, Curitiba, PR, Brazil\\
Email: alceu@ppgia.pucpr.br}
}

\maketitle

\begin{abstract}
Automatic facial expression recognition is an important research area in the emotion recognition and computer vision. Applications can be found in several domains such as medical treatment, driver fatigue surveillance, sociable robotics, and several other human-computer interaction systems. Therefore, it is crucial that the machine should be able to recognize the emotional state of the user with high accuracy. In recent years, deep neural networks have been used with great success in recognizing emotions. In this paper, we present a new model for continuous emotion recognition based on facial expression recognition by using an unsupervised learning approach based on transfer learning and autoencoders. The proposed approach also includes preprocessing and post-processing techniques which contribute favorably to improving the performance of predicting the concordance correlation coefficient for arousal and valence dimensions. Experimental results for predicting spontaneous and natural emotions on the RECOLA 2016 dataset have shown that the proposed approach based on visual information can achieve CCCs of 0.516 and 0.264 for valence and arousal, respectively.
\end{abstract}

\begin{IEEEkeywords}
Deep learning, Unsupervised Learning, Representation learning, Facial Expression Recognition
\end{IEEEkeywords}

%%%%%%%%%%%%%%%%%%%%%%%%%%%%%%%%%%%%%%%%%%%%%%%%%%%%%%
\section{Introduction}
%%%%%%%%%%%%%%%%%%%%%%%%%%%%%%%%%%%%%%%%%%%%%%%%%%%%%%
The visual recognition of emotional states usually involves analyzing a person’s facial expression, body language, or speech signals. Facial expressions contain abundant and valuable information about the emotion and thought of human beings. Facial expressions naturally transmit emotions even if a subject wants to mask his/her emotions. Several researchers suggest that there are emotional strokes produced by the brain and shown involuntarily by our corps through the face \cite{b1}. Emotions are an important process for human-to-human communication and social contact. Thus, emotions need to be considered to achieve better human-machine interaction.

According to theories in psychology research \cite{b1,b2}, there are three emotion theories to model the emotion state: discrete theory, appraisal theory and dimensional theory. The first one, the discrete theory claims that there exists a small number of discrete emotions (i.e., anger, disgust, happiness, neutral, sadness, fear, and surprise) that are inherent in our brain and recognized universally \cite{b3}. 

\begin{figure}[htbp]
		\centering
		\includegraphics[width=0.4\textwidth]{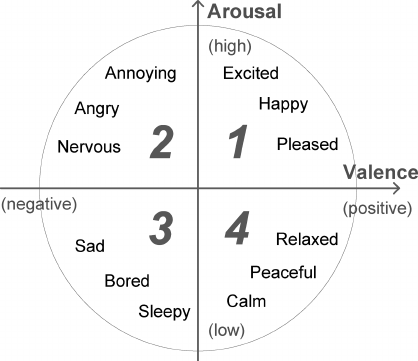}
	\caption{Valence-Arousal 2D dimension plane \cite{b40}.}
	\label{fig:Image/Valence-Arousal 2D dimension plane.}
\end{figure}

That one has been largely adopted in research on emotion recognition. However, it has some drawbacks as it does not take into consideration people who exhibit non-basic, subtle and complex emotions like depression. It results that these basic discrete classes may not reflect the complexity of the emotional state expressed by humans. As a result, the appraisal theory has been introduced. 
This is a theory where emotions are generated through continuous, recursive subjective evaluation of both our own internal state and the state of the outside world \cite{b3}. Nonetheless, the appraisal theory is still an open research problem on how to use it for automatic measurement of emotional state. For the dimensional theory, the emotional state  considers a point in a continuous space. This third theory can model the subtle, complicated and continuous emotional state. It models emotions using two independent dimensions, i.e.arousal  (relaxed  vs.  aroused)  and  valence  (pleasant  vs. unpleasant) as shown in Fig.~\ref{fig:Image/Valence-Arousal 2D dimension plane.}. The valence dimension refers to how positive or negative emotion is, and range from unpleasant to pleasant. The arousal dimension refers to how excited or apathetic emotion is, ranging from sleepiness or boredom to frantic excitement \cite{b4}.

The typical approach is to take every single data as a single unit (e.g., a frame of a video sequence) independently. It can be made as a standard regression problem for every frame using the so-called static (frame-based) regressors. Many researches have been scrutinized by predicting emotion in continuous dimensional space from the recognition of discrete emotion categories. However, emotion recognition is a challenging task because human emotions lack temporal boundaries. Moreover, each individual expresses and perceive emotions in different ways. In addition, one utterance may contain more than one emotion.

Several deep learning architectures such as convolutional neural networks (CNNs), autoencoder (AE), memory enhanced neural network models such as long short-term memory (LSTM) models,  have recently been used successfully for emotion recognition. Traditionally facial expression recognition consists of feature extraction utilizing handcrafted representations such as Local Binary Pattern (LBP) \cite{b5,b33,b34,b35,b36}, Histogram of Oriented Gradients (HOG) \cite{b7,b37}, Scale Invariant Feature Transform (SIFT) \cite{b6},  Gabor wavelet coefficients \cite{b27,b28,b29,b30,b31}, Haar features \cite{b29,b32}, 3D shape parameters \cite{b38} and then predict the emotion from these extracted features. Shan et al. \cite{b5} formulated a boosted-LBP feature and combined it with a support vector machine (SVM) classifier. Berretti et al. \cite{b6} computed the SIFT descriptor on 3D facial landmarks of depth images and used SVM for the classification. Albiol et al. \cite{b7} proposed a HOG descriptor-based EBGM algorithm that is more robust to changes in illumination, rotation, small displacements and to the higher accuracy of the face graphs obtained compared to classical Gabor–EBGM ones.

A number of studies in the literature have focused on predicting emotion from face detection using deep neural networks (DNN). Zhao et al. \cite{b8} combined deep belief networks (DBN) and multi-layer perceptron (MLP) for facial expression recognition. Mostafa et al. \cite{b9} used recurrent neural networks (RNN) to study emotion recognition from facial features extracted. A large majority of these scientific studies had been carried out using the handcrafted features. Despite the fact that these approaches reported good accuracy for the prediction, the handcrafted feature has its inherent drawbacks; either unintended features that do not benefit classification may be included or important features that have a great influence on the classification may get omitted. This is because these features are “crafted” by human experts, and the experts may not be able to consider all possible cases and include them in feature vectors.

With the recent success achieved in deep learning, a trend in machine learning has emerged towards deriving a representation directly from the raw input signal. Such a trend is motivated by the fact that CNNs learn representation and discriminant functions through iterative weight updated by backpropagation and error optimization. Therefore, CNNs could include critical and unforeseen features that humans hardly come up with and hence contribute to improving the performance. CNNs have been employed in many works but oftentimes, they require a high number of convolutional layers to learn a good representation due to the high complexity of facial expression images. The disadvantage of increasing network depth is the complexity of the network as the training time, which can grow significantly with each additional layer. Furthermore, increasing network complexity requires more training data and it makes it more difficult to find the best network configuration as well as the best initialization parameters. %for deep networks in a supervised learning approach is always challenging and it requires a large number of attempts to move towards the best possible recognition performance. 

In this paper, we introduce unsupervised feature learning to predict the emotional state in an end-to-end approach. We aim to learn good representations in order to build a compact continuous emotion recognition model with a reduced number of parameters that produce a good prediction. We propose a convolutional autoencoder (CAE) architecture, which learns good representations from facial images while reducing the high dimensionality. The encoder is used to compress the data and the decoder is used to reproduce the original image. %Therefore, autoencoders may be used for data compression. Compression logic is data-specific, meaning it is learned from data rather than predefined compression algorithms such as JPEG, MP3, and so on.%
%It enforces a meaningful representation.
%Features are extracted from facial information using a CAE architecture. 
The representation learnt by the CAE is used to train a support vector regressor (SVR) to predict the affective state of individuals. In this architecture we did not take into consideration the temporal aspect of the raw signals.

The main contributions of this paper are: (i) reduction of the dimensionality; (ii) we only used raw images without handcrafted; (iii) a representation learned from unlabeled raw data that is comparable to the state-of-the-art and that achieved CCCs that are comparable to the state-of-the-art as well. 
%{\color{red} ALEKOE -- These contributions are very weak and are not really true. Try to rewrite them}. 

The structure of this paper is as follow. Section~\ref{sec:rel} provide the most recent studies on emotion recognition from facial expression. Section~\ref{sec:prop} introduces our model. In Section~\ref{sec:data}, we describe the dataset used in this study. We present our results in Section~\ref{sec:exp}. Conclusions and perspectives of future work are presented in the last section.

%%%%%%%%%%%%%%%%%%%%%%%%%%%%%%%%%%%%%%%%%%%%%%%%%
\section{Related Work}
\label{sec:rel}
%%%%%%%%%%%%%%%%%%%%%%%%%%%%%%%%%%%%%%%%%%%%%%%%
\begin{figure*}[htbp!]
	\centering
	\includegraphics[width=0.9\textwidth]{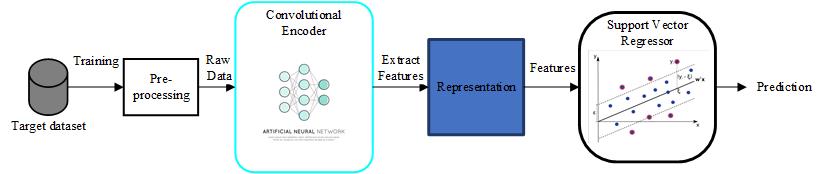}
	\caption{Overview architecture}
	\label{fig:Image/overview_of_Process_Static_Image.jpg}
\end{figure*}

A number of studies have been proposed to model facial expression recognition (FER) from the raw image with DNNs. Tang \cite{b10} used L2-SVM objective function to train DNNs for classification. Lower layer weights are learned by backpropagating the gradients from the top layer linear SVM by differentiating the L2-SVM objective function with respect to the activation of the penultimate layer. Moreover, Liu et al. \cite{b11} proposed a 3D-CNN and deformable action part constraints in order to locate facial action parts and learn part-based features for emotion recognition. In the same vein, Liu et al. \cite{b12} extracted image-level features with pre-trained Caffe CNN models. In addition, Yu and Zhang \cite{b13} proved that the random initialization of neural networks allowed to vary network parameters and also renders the classification ability of diverse networks. Because of that, the ensemble technique usually shows concrete performance improvement. Furthermore, Kahou et al.\cite{b14} proposed an approach that combines multiple DNNs for different data modalities such as facial images, audio, bag of mouth features with CNN, deep restricted Boltzmann machine and the output of such modalities are averaged to take a final decision. Liu et al. \cite{b26} presented a boosted DBN to combine feature learning/strengthen, feature selection and classifier construction in a unified framework. Features are fine-tuned and jointly selected to form a strong classifier that can learn highly complex features from facial images and more importantly, the discriminative capabilities of selected features are strengthened iteratively according to their relative importance to the strong classifier. Mollahosseini et al. \cite{b39} proposed a single component architecture made up of two convolutional layers each, followed by max pooling and four inception layers. The inception layers increase the depth and width of the network while keeping the computational budget constant. 

So far, plenty of papers for FER used CNN and their structures and image preprocessing techniques were all different. Mostly, they used a supervised approach where labeling is expensive, then it will be difficult to handle a large dataset. This is a limitation because nowadays a lot of unlabeled data are created continuously. It is imperative that automatic FER must deal with this case and take advantage of it. The proposed approach differs from the previous ones in a way that it has the ability to handle large datasets with the unsupervised approach, to learn the inherent relevant features without using explicitly provided labels and then predict emotional state with high accuracy.

%%%%%%%%%%%%%%%%%%%%%%%%%%%%%%%%%%%%%%%%%%%%%%%%%%%%%%%%
\section{Proposed Approach}
\label{sec:prop}
%%%%%%%%%%%%%%%%%%%%%%%%%%%%%%%%%%%%%%%%%%%%%%%%%%%%%%%%
In this section, we describe the overall architecture of the proposed model, which is made up of three parts, as shown in Fig.~\ref{fig:Image/overview_of_Process_Static_Image.jpg}. A key component of our model is the convolution operation and autoencoder. Traditionally, most studies on facial expression recognition are based on handcrafted features. However, after the success of DNNs, many works on facial expression recognition are now based on supervised approaches for representation learning using CNNs.

In contrast to previous works in facial expression recognition, the proposed approach starts with supervised learning on a source dataset, transfer learning to initialize the convolution layers of a CAE, unsupervised learning to learn a meaningful representation on a target dataset, and again, a supervised approach to train a regression model to predict continuous emotions.

\begin{figure}[htbp!]
	\centering
	\includegraphics[width=0.5\textwidth]{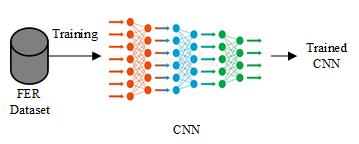}
	\caption{Pre-training a CNN in a source dataset}
	\label{fig:Image/Transfer_Learning_model_Video_V5.png}
\end{figure}

\begin{table}[htbp]
\centering
\caption{Pre-training CNN Architecture}
\label{tab:Pre-training CNN Architecture}
%\resizebox{\textwidth}{!}{%
\begin{tabular}{|l|c|c|l|}
\hline
\textbf{\begin{tabular}[c]{@{}l@{}}Layers\\ Type\end{tabular}} &
  \textbf{\begin{tabular}[c]{@{}l@{}}Filter\\ Dimension\end{tabular}} &
  \textbf{\begin{tabular}[c]{@{}l@{}}Kernel\\ Size\end{tabular}} &
  \textbf{Activation} \\ \hline
Conv2D             & 64  & 3$\times$3 & reLu    \\ \hline
BatchNormalization & -   & -   & -       \\ \hline
Conv2D             & 64  & 3$\times$3 & tanh    \\ \hline
Max Pool           & -   & 2$\times$2 & -       \\ \hline
BatchNormalization & -   & -   & -       \\ \hline
Conv2D             & 128 & 2$\times$2 & reLu    \\ \hline
Max Pool           & -   & 2$\times$2 &         \\ \hline
Flatten            & -   & -   & -       \\ \hline
Fully Connected    & 100 & -   & tanh    \\ \hline
Dropout(0.5)       & -   & -   & -       \\ \hline
Fully Connected    & 50  & -   & reLu    \\ \hline
Fully Connected    & 10  & -   & tanh    \\ \hline
Fully Connected    & 7   & -   & softmax \\ \hline
\end{tabular}%
%}
\end{table}

In the first stage, we begin with transfer learning technique that allows us to import information from another model to jump start the development process on a new or similar task. The key concept is to use FER dataset, which has been used in ICMLW2013\footnote{30th International Conference on Machine Learning - Workshop on Representational Learning} \cite{b10} to recognize discrete emotions in pictures. This dataset provides a large number of facial images with emotional content to train a CNN. Once we finish training the CNN model on the FER dataset, we use such a pre-trained CNN to initialize the CAE, as shown in Fig.~\ref{fig:Image/Transfer_Learning_model_Video_V5.png}.

\begin{figure*}[htbp!]
	\centering
	\includegraphics[width=0.9\textwidth]{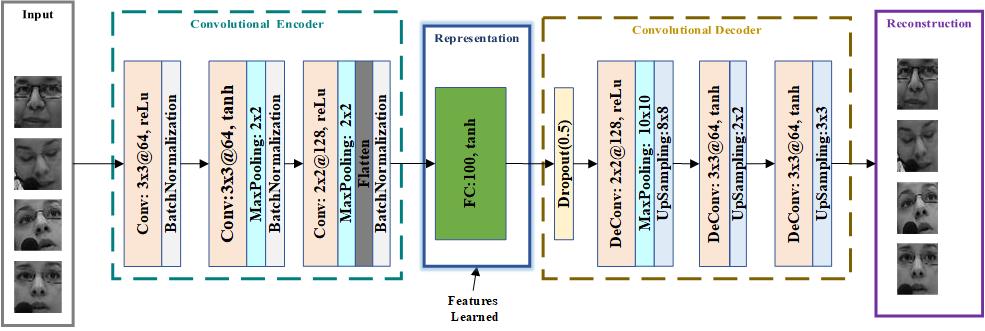}
	\caption{Proposed Convolutional Autoencoder (CAE)}
	\label{fig:Image/Video_CNN_Autoencoder_V7_StaticImages.jpg}
\end{figure*}

The proposed pre-training CNN shown in Table \ref{tab:Pre-training CNN Architecture} is inspired by the architecture proposed by Sun et al.~\cite{b15} which achieved 67.8\% of accuracy on the test set of the FER dataset. In order to enhance the learned representation and reduce the number of parameters of the network, we changed on it the number of convolutional layers and fully connect layers.
In addition, to have the best trade-off between the complexity and the amount of data available. The pre-trained model has three convolutional layers and four fully connected layers as shown in Table \ref{tab:Pre-training CNN Architecture}. The first convolutional layer filters the input patch with 64 kernels of size 3$\times$3, reLu activation followed by a batch normalization layer. The second convolutional layer takes as input the response-normalized of the first convolutional layer and filters it with 64 kernels of size 3$\times$3$\times$64, \textit{tanh} activation with a maxpooling. The third convolutional layer has 128 kernels of size 2$\times$2$\times$128, reLu activation connected to the (normalized, pooled) outputs of the second convolutional layer. The first, second, third and fourth fully connected (FC) layers have 100, 50, 10, and 7 neurons respectively. %For last, layer’s output is used as the regression value of the whole network.
The CNN is trained up to 500 epochs using a categorical crossentropy loss function with Adam optimizer.

In the next step, we have the unsupervised approach for representation learning. A CAE is a convolutional neural network architecture for unsupervised learning that is trained to reproduce its input image in the output layer. An image is passed through an encoder, which is a CNN that produces a low-dimensional representation of the image. The decoder, which is another CNN,  takes this compressed representation and try to reconstruct the original image. The architecture of the proposed CAE is shown in Fig.~\ref{fig:Image/Video_CNN_Autoencoder_V7_StaticImages.jpg}. The convolutional encoder has three convolutional layers and one fully connected layer. The first convolutional layer filters the input patch with 64 kernels of size 3$\times$3, reLu activation followed by a batch normalization layer. The second convolutional layer takes as input the response-normalized of the first convolutional layer and filters it with 64 kernels of size 3$\times$3$\times$64, tanh activation with a maxpooling. The third convolutional layer has 128 kernels of size 2$\times$2$\times$128, reLu activation connected to the (normalized, pooled) outputs of the second convolutional layer. The fully connected layer has a certain number of neurons. The decoder network also has 3 convolutional layers. The first convolutional layer filters the output of the fully connected layer patch with 128 kernels of size 2$\times$2, reLu activation followed by a max pooling layer and an upsampling layer. The second convolutional layer takes as input the output of the first convolutional layer and filters it with 64 kernels of size 3$\times$3$\times$64, \textit{tanh} activation with an upsampling layer. The third convolutional layer has 64 kernels of size 3$\times$3$\times$64, reLu activation connected to the outputs of the second convolutional layer.

In the final step, we have the supervised approach for regression that uses as input features the representation learned at the fully connected layer (encoder layer) of the autoencoder. An SVR is trained with the features generated by the CAE to predict continuous emotions. We follow the strategy that has been used in AVEC 2016\footnote{Audio-Visual + Emotion Recognition Challenge} \cite{b21} to predict the arousal as well as the valence. We use grid search to find the best combination of the complexity parameter $C$ and epsilon $\epsilon$ that maximize a performance measure.

%%%%%%%%%%%%%%%%%%%%%%%%%%%%%%%%%%%%%%%%%%%%%%%%%%%%%%%%
\subsection{Post-Processing}
%%%%%%%%%%%%%%%%%%%%%%%%%%%%%%%%%%%%%%%%%%%%%%%%%%%%%%%%
As our architecture does not take into consideration the correlation between the sequence of images, the predictions obtained may have some noises. To handle this issue we apply techniques such as median filter, scaling, and centering, which allow us to improve the performance. 

Median filtering is usually used to reduce potential noise from the images. It turns out that it does a pretty good job of preserving edges in an image. It makes our prediction smooth by reducing its high-frequency components by filtering the 1D output array with a size of window between 0.04 and 20 seconds. 

The scaling factor $\beta$ is the ratio obtained by the gold standard $(GS)$ and the prediction$(Pr)$ as shown in \eqref{eq:scalingfactor} over the training set. The prediction on the development set is multiplied by this factor $\beta$ with the purpose to rescale the output as shown in \eqref{eq:multiplyfactor}. 

\begin{equation}
    \beta_{tr} = \frac{GS_{tr}}{Pr_{tr}}
    \label{eq:scalingfactor}
\end{equation}

\begin{equation}
  {Pr^{'}_{dev}}   =  \beta_{tr} \cdot {Pr_{dev}}
    \label{eq:multiplyfactor}
\end{equation}

The centering technique entails just subtracting the mean of the predictions by the prediction as shown in \eqref{eq:centering}, where $y$ is the prediction and $y^{'}$ is the corrected value by subtracting the mean over the gold standard.

\begin{equation}
  {y^{'}}   =  {y} \text{-} \overline{y}_{GS}
    \label{eq:centering}
\end{equation}

%%%%%%%%%%%%%%%%%%%%%%%%%%%%%%%%%%%%%%%%%%%
\section{Datasets} 
\label{sec:data}
%%%%%%%%%%%%%%%%%%%%%%%%%%%%%%%%%%%%%%%%%%%
In this section we present the source dataset used for pre-training the convolutional layers of a CNN and the target dataset used for representation learning and final prediction. Furthermore, we also present the preprocessing techniques used to detect and align face images within video frames.

%%%%%%%%%%%%%%%%%%%%%%%%%%%%%%%%%%%%%%%%%%%
\subsection{FER Dataset}
%%%%%%%%%%%%%%%%%%%%%%%%%%%%%%%%%%%%%%%%%%%
The Facial Expression Recognition dataset known as FER has been created by P. L. Carrier and A. Courville and it is freely available. We used FER dataset as the starting point in the pre-training stage for the proposed model shown in Fig.~\ref{tab:Pre-training CNN Architecture}. The FER dataset is made up of grayscale images of 48$\times$48 pixels that comprise seven acted emotions (disgust, anger, fear, joy, saddens, surprise, neutral) and it is split into 28,709 images for training, 3,589 for validation and 3,589 for test. Besides, it offers individual images that can be correlated if they have the same labeled emotion.

%%%%%%%%%%%%%%%%%%%%%%%%%%%%%%%%%%%%%%%%
\subsection{RECOLA Dataset}
%%%%%%%%%%%%%%%%%%%%%%%%%%%%%%%%%%%%%%%
To evaluate our architecture, we use the RE-mote COLlaborative and Affective (RECOLA) dataset introduced by Ringeval et al. \cite{b18} to study socio-affective behaviors from multimodal data in the context of remote collaborative work for the development of computer-mediated communication tools\cite{b19}. A subset of the dataset was used in the Audio/Visual Emotion Challenge and Workshop (AVEC) 2015 and 2016 challenges \cite{b17, b21}. However in this study, we use the subset of the dataset used for AVEC 2016 \cite{b21} as we do not have the full contents. It contains four modalities that are audio, video, electrocardiogram (ECG) and electro-dermal activity (EDA). The dataset is split equally in three partitions–train (9 subjects), validation (9 subjects) and test (9 subjects)–by stratifying (i.e., balancing) the gender and the age of the speakers. The labels of the RECOLA are re-sampled at a constant frame rate of 40 ms. In addition, we do not have the test set. We ensured that no validation data were used for unsupervised feature learning. The CAE has been trained with all available unlabeled video data.

%%%%%%%%%%%%%%%%%%%%%%%%%%%%%%%%%%%%%%%%%%%%
\subsection{Face Detection and Alignment} 
%%%%%%%%%%%%%%%%%%%%%%%%%%%%%%%%%%%%%%%%%%%%
On facial expression recognition several obstacles appear in our path to achieve a suitable prediction. One of them being the fact that humans as unpredictable entities are in a constant movement even in a face to face conversation and because of this, sometimes, the subject does not look directly into the camera. Other issues arise like a delay on the annotated labels that is imposed by the annotator and the absence of bounding box coordinates for all the frames. Therefore, we tried different strategies. We started with the dropping frames issue. Over the RECOLA dataset, a certain number of frames do not have the bounding box coordinates to extract the face. Even with our own-implemented face-detector we cannot extract for all the frames, the section over the image that contains the face. Because of this, we tried to preserve all the dataset by using the entire image without the bounding box. The frame quality selection to filter detected face are far from being frontal face images and the delay compensation to realign labels and frames to compensate for the reaction lag of annotators. By doing so, we reduced the dataset size, which is actually not a good option for our approach, because the unsupervised algorithms perform well with a large amount of data. Instead of dropping the blank frames or frames where faces are not well detected, we decided to change them by other frames well detected. It is a kind of data augmentation strategy whereby the frames without the bounding box have been slightly changed to detect the face of participants.

%%%%%%%%%%%%%%%%%%%%%%%%%%%%%%%%%%%%%%%%%%
\section{Experiments and Results}
\label{sec:exp}
%%%%%%%%%%%%%%%%%%%%%%%%%%%%%%%%%%%%%%%%%%
This section presents the metrics used and the experiments undertaken to evaluate the proposed approach. The experimental results are analyzed and compared to previous works.

%%%%%%%%%%%%%%%%%%%%%%%%%%%%%%%%%%%%%%%%%%%%%%%%%%%%%%%%
\subsection{Metrics}
%%%%%%%%%%%%%%%%%%%%%%%%%%%%%%%%%%%%%%%%%%%%%%%%%%%%%%%%

Concordance Correlation Coefficient (CCC) \cite{b16} %\cite{ConcordanceCorrelationCoefficient}
is calculated as the evaluation metric for this challenge. It combines Pearson’s Correlation Coefficient (CC) with the square difference between the mean of the two compared time-series as denoted in \eqref{equation CCC}.
\begin{equation}
    \rho _{c} = \frac{2\rho\sigma _{x}\sigma _{y}}{\sigma _{x}^{2} + \sigma _{y}^{2} + (\mu _{x} - \mu _{y})^{2}}
    \label{equation CCC}
\end{equation}

\noindent where $\rho$ is the Pearson correlation coefficient between two time-series (e. g., prediction and gold standard), $\sigma _{x}^{2}$ and $\sigma _{y}^{2}$ the variance of each time-series and $\mu _{x}$ and $\mu _{}$ the mean value of each. As a result, predictions that are well correlated with the gold standard but shifted in value are penalised in proportion to the deviation.

CCC will help to evaluate emotion recognition in terms of continuous time and continuous valued dimensional affect into two dimensions: arousal and valence \cite{b17}. The problem of dimensional emotion recognition can thus be posed as a regression problem through two dimensions.

%%%%%%%%%%%%%%%%%%%%%%%%%%%%%%%%%%%%%%%%%%
 \subsection{Experimental Setup}
%%%%%%%%%%%%%%%%%%%%%%%%%%%%%%%%%%%%%%%%%%
For raw signal, we cropped faces of the subject’s video to have the images with the size 48$\times$48. The image size 48$\times$48 is used to reduce the computation complexity and because the pre-trained model used the FER dataset, which consists of 48$\times$48 pixel grayscale images of faces.
To train the proposed model, we initialized the network with the pre-trained weights from the FER dataset. We used the Adam optimization method MSE as loss function and a fixed learning rate of $10^{-5}$  throughout all experiments. We tried different batch sizes, learning rates and epochs in order to determine the best setup for the model training. We tried different dimensions for the encoded layer. For regularization of the network, we also used dropout with $p$ = 0.25 for all layers. This step is important as our models have a large number of parameters and not regularizing the network makes it prone to overfitting on the training data. 
 
We have carried out different experiments by freezing the different convolutional layers (CL) and fine-tune (training) just the fully connected (FC) as shown in Table~\ref{tab: CCC_Arousal_Valence_freezing_Conv_Video}. Alternately, we unfreeze one convolutional layer per training session from the deeper to the first convolution layer and retrain the network with RECOLA dataset.

\begin{table}[htbp]
\caption{CCC Scores for the Arousal and Valence dimension by freezing different numbers
of convolutional layers.}
\centering
\begin{tabular}{|l|c|c|c|c|}
\hline
\multicolumn{2}{|l|}{} & \multicolumn{3}{c|}{\textbf{CCC}} \\ \hline
\multicolumn{2}{|l|}{\textbf{Dimension}} & 0 Conv-Frozen & 2 Conv-Frozen & 1 Conv-Frozen \\ \hline
\multicolumn{2}{|l|}{Valence} & 0.397 & 0.399 & \textbf{0.516} \\ \hline
\multicolumn{2}{|l|}{Arousal} & 0.027 & 0.035  & \textbf{0.264} \\ \hline
\end{tabular}
%}
\label{tab: CCC_Arousal_Valence_freezing_Conv_Video}
\end{table}

We noticed that unfreezing only one CL of the deepest layer gives the best result. By the way, when all the CLs are frozen, the model did not train properly, meaning that the value of the loss function did not decrease significantly. In the end, a chain of post-processing methods is applied, namely, median filtering (size of window was between 0.04s and 20s) \cite{b20}, centering (by finding the ground truth’s and the prediction’s bias) \cite{b22}, scaling (with scaling factor given by the ration between the standard deviation of the ground truth and the prediction computed at the training set) \cite{b22} and time-shifting (forward in time with values between 0.04s and 10s) \cite{b23}. Any of these post-processing techniques were kept when we have observed an improvement in the CCC.

%%%%%%%%%%%%%%%%%%%%%%%%%%%%%%%%%%%%%%%%%%%%%%%
\subsection{Results}
%%%%%%%%%%%%%%%%%%%%%%%%%%%%%%%%%%%%%%%%%%%%%%%
We realized that when the encoder layer has a small dimension, the CCC score is very low as shown in Table~\ref{tab:CCC for SVR trained on CAE features for arousal and valence dimensions}. In other words, the CCC score increases when the encoder layer size increases, but there is an upper limit. This is due to the fact that by increasing the encoder layer the CAE is able to learn more relevant representations. Nonetheless, we also found out from some dimensions specifically 1,000 neurons for the encoder layer that the CCC seems not to increase. That can be explained by the fact that the CAE does not find novel relevant features and this behavior is related to the size of the training set. By augmenting the number of training samples, probably the CAE will probably continue to capture the features.

\begin{table}[htbp]
\caption{CCC Scores for SVR trained on CAE features for arousal and valence dimensions with delay compensation of 40 and 30 frames.}
\centering
\begin{tabular}{|l|c|c|c|}
\hline
\textbf{Dimension} & \textbf{Encoder Layer} & \textbf{Delay} & \textbf{CCC} \\ \hline
Valence & 100 & 40 & 0.197 \\ \hline
Valence & 500 & 40 & 0.324 \\ \hline
Valence & 700 & 40 & 0.361 \\ \hline
Valence & 900 & 40 & \textbf{0.516} \\ \hline
Valence & 1000 & 40 & 0.365 \\ \hline
Valence & 100 & 30 & 0.195 \\ \hline
Valence & 500 & 30 & 0.384 \\ \hline
Valence & 700 & 30 & 0.392 \\ \hline
Valence & 900 & 30 & 0.498 \\ \hline
Valence & 1000 & 30 & 0.395 \\ \hline
Arousal & 100 & 40 & 0.018 \\ \hline
Arousal & 500 & 40 & 0.071 \\ \hline
Arousal & 700 & 40 & 0.151\\ \hline
Arousal & 900 & 40 & \textbf{0.264} \\ \hline
Arousal & 1000 & 40 & 0.119 \\ \hline
Arousal & 100 & 30 & 0.031 \\ \hline
Arousal & 500 & 30 & 0.092 \\ \hline
Arousal & 700 & 30 & 0.162 \\ \hline
Arousal & 900 & 30 & 0.257 \\ \hline
Arousal & 1000 & 30 & 0.114 \\ \hline
\end{tabular}
\label{tab:CCC for SVR trained on CAE features for arousal and valence dimensions}
\end{table}

We compare the performance achieved by our method against the current state-of-the-art for the RECOLA dataset as shown in Table~\ref{tab:Performance_comparison_between_the_proposed_methodCAE_And_StateofArts}. Most of them these results have been submitted to the AVEC2016 challenge which used a subset of RECOLA dataset encompassing only 27 participants. % We should note that Tzirakis et al. \cite{b24} also used DNNs but they used the full dataset with 46 participants, data augmentation and take into consideration the temporal correlation between frames.
We observed that the results obtained by Tzirakis et al. \cite{b24} are slightly higher than the proposed approach because of the dataset size. They used a dataset with 46 participants and the temporal context whereas we used 27 participants. The prediction of the valence dimension of our model outperforms  Han et al. \cite{b25}  even though the prediction of the arousal dimension remains the same. In comparison with AVEC2016 \cite{b21}, the prediction of the valence dimensions of our model outperforms the appearance features and it is slightly higher than the geometric features. The prediction of the arousal dimension of our model is slightly less than those of appearance and geometric features.

\begin{table}[htbp]
\caption{Performance comparison between the proposed method (CAE) and other state-of-the-art methods}
\centering
\begin{tabular}{|l|l|c|c|}
\hline
\textbf{Predictors} & \textbf{Features} & \textbf{Valence} & \textbf{Arousal} \\ \hline
Baseline \cite{b24}  
& Raw signal & 0.620 & 0.435 \\ \hline
AVEC 2016 \cite{b21} 
& Appearance & 0.486 & 0.343 \\ \hline
AVEC 2016 \cite{b21}  
& Geometric & 0.507 & 0.272 \\ \hline
Han et al. \cite{b25} 
& Mixed & 0.265 & 0.394 \\ \hline
Proposed & Raw signal & \textbf{0.516} & \textbf{0.264} \\ \hline
\end{tabular}

\label{tab:Performance_comparison_between_the_proposed_methodCAE_And_StateofArts}
\end{table}

%%%%%%%%%%%%%%%%%%%%%%%%%%%%%%%%%%%%%%%%%%%%%%%%
\section{Conclusion}
%%%%%%%%%%%%%%%%%%%%%%%%%%%%%%%%%%%%%%%%%%%%%%%%
A large number of existing works conducted expression recognition tasks based on a static image without considering the temporal context. In this paper, we propose an unsupervised approach based on CAE for recognizing facial expressions from static images of faces. We preprocessed the data. We started by extracting features from the convolutional filter without the labels. This approach greatly reduced the feature dimension and computation, making the recognition system more efficient. The results obtained outperform some results from the literature and competitive with the baseline \cite{b21}.  
Future work can extend the framework to capture temporal dependence over the video data.

%{\color{red}ALEKOE --- Too short. Improve it}

\balance

\vspace{12pt}
%\color{red}IEEE conference templates contain guidance text for composing and formatting conference papers. Please ensure that all template text is removed from your conference paper prior to submission to the conference. Failure to remove the template text from your paper may result in your paper not being published.

\end{document}